\documentclass{article} 
\usepackage{iclr2016_conference,times}
\usepackage{hyperref}
\usepackage{url}
\usepackage{bm}
\usepackage{array}
\usepackage{graphicx}
\usepackage{amsmath}
\usepackage{bbm}
\usepackage{amssymb}
\usepackage{enumitem}
\usepackage{subfig}
\newcolumntype{L}[1]{>{\raggedright\let\newline\\\arraybackslash\hspace{0pt}}m{#1}}
\newcolumntype{C}[1]{>{\centering\let\newline\\\arraybackslash\hspace{0pt}}m{#1}}
\newcolumntype{R}[1]{>{\raggedleft\let\newline\\\arraybackslash\hspace{0pt}}m{#1}}

\newcommand{\be}{\begin{equation}}
\newcommand{\ee}{\end{equation}}
\newcommand{\bea}{\begin{eqnarray}}
\newcommand{\eea}{\end{eqnarray}}

\newcommand\norm[1]{\left\lVert#1\right\rVert}

\title{Improving performance of recurrent neural network with relu nonlinearity}

\author{Sachin S. Talathi \& Aniket Vartak \\
Qualcomm Research\\
San Diego, CA 92121, USA \\
\texttt{\{stalathi,avartak\}@qti.qualcomm.com} \\
}

\begin{document}

\maketitle

\begin{abstract}
In recent years significant progress has been made in successfully training recurrent neural networks (RNNs) on sequence learning problems involving long range temporal dependencies. The progress has been made on three fronts: (a) Algorithmic improvements involving sophisticated optimization techniques, (b) network design involving complex hidden layer nodes and specialized recurrent layer connections and (c) weight initialization methods. In this paper, we focus on recently proposed weight initialization with identity matrix for the recurrent weights in a RNN. This initialization is specifically proposed for hidden nodes with Rectified Linear Unit (ReLU) non linearity. We offer a simple dynamical systems perspective on weight initialization process, which allows us to propose a modified weight initialization strategy. We show that this initialization technique leads to successfully training RNNs composed of ReLUs. We demonstrate that our proposal produces comparable or better solution for three toy problems involving long range temporal structure: the addition problem, the multiplication problem and the MNIST classification problem using sequence of pixels. In addition, we present results for a benchmark action recognition problem.
\end{abstract}

\section{Introduction}
Recurrent neural networks (RNNs) are neural networks with cyclical connections between the hidden nodes. These cyclical connections offer RNNs the ability to encode memory and as such these networks, if successfully trained, are suitable for sequence learning applications. Traditionally, training RNNs using stochastic gradient descent methods such as back-propagation through time (BPTT) \citep{Rumelhart_1986} have been riddled with difficulty. Early attempts suffered from the so-called vanishing gradient or exploding gradient problems resulting in difficulties to learn long-range temporal dependencies \citep{Hochreiter_2001}.

Several methods have been proposed to overcome the difficulty in training RNNs. These methods broadly fall into three categories: (i) Algorithmic improvements involving sophisticated optimization techniques. The Hessian-Free (HF) optimization method \citep{Martens_2010,Martens_2011} for training RNNs falls under this category. HF method works by taking large steps in the weight space in directions of low gradient and curvature. As noted by the authors, it is the ability of HF to identify and pursue these directions, which allow HF to fully optimize deep neural networks, such as the RNNs unfolded in time. (ii) Network design, the most successful technique to date under this category is the Long Short Term Memory (LSTM) RNN \citep{Hochreiter_1997}. The LSTM is a standard RNN in which the monotonic nonlinearity such as the tanh or sigmoid is replaced with a memory unit that can efficiently store continuous values and transmit information over a long temporal range.  Each memory unit in the LSTM has fixed linear dynamics with a feedback loop of weight one. As a result the error signal neither decays nor explodes as it back propagates through the memory unit. 
The memory cell content is modulated by various gates that allow the LSTM unit to store and retrieve relevant sequence information. The addition of various gates, however add to the complexity of LSTM unit. A typical LSTM network has four times more trainable parameters than an equivalent RNN network comprised of monotonic nonlinearity \citep{Klaus_2015}. LSTM RNNs and variants thereof \citep{Cho_2014} have produced impressive results on several sequence learning tasks including handwritting recognition \citep{Graves_2013}, speech recognition \citep{Graves_2014}, machine translation \citep{Cho_2014,Bahdanau_2014}, image captioning \citep{Kiros_2014}, predicting output of simple computer programs \citep{Zaremba_2014} and action recognition from video clips \citep{Ng_2015}.  More recently, a modification to the standard RNN architecture, the clockwork RNN (CW-RNN) has been proposed \citep{Koutnik_2014}. The long range dependency problem of vanishing and exploding gradients in the standard RNN is solved in CW-RNN by dividing the hidden layer of the CW-RNN into separate modules, each running at different clock speeds. This allows CW-RNNs to efficiently learn different time scales inherent in complex-signals. (iii) Weight initialization, in a recent paper by Quoc et al, \citep{Quoc_2015}, the authors propose a simpler solution to the long-range dependence problem with RNNs composed of rectified linear units (ReLU)s. They use the Identity matrix to initialize the recurrent weight matrix.They call the resulting RNN, the Identity-RNN (IRNN). The basic idea underlying IRNN is that in the absence of any input, a RNN composed of ReLUs and initialized with identity weight matrix just stays the same indefinitely. Experimental results with IRNN on several benchmark problems with long-range temporal structure are comparable to the those produced by LSTM RNNs \citep{Quoc_2015}.

Given our desire to design low-complexity, low memory foot-print RNN models for applications on mobile computing platform, we were particularly interested in the encouraging results produced by IRNN networks. In this paper, we therefore focus on IRNNs and investigate the significance of identity weight initialization in IRNNs from a dynamical systems perspective.  We demonstrate that in absence of any input and with identity weight initialization the hidden nodes exhibit neutrally stable fixed point dynamics. While this observation produces the effect that the hidden node dynamics stays the same indefinitely in absence of any input, the node dynamics is also extremely sensitive to input perturbations. We hypothesize that this sensitivity of hidden node dynamics to input may influence the success of RNN for sequence learning tasks. Motivated by this hypothesis we propose a new weight initialization matrix, the normalized-positive definite weight matrix for RNNs comprised of ReLU hidden units. The resulting RNN is refered to as the np-RNN. We show that on several of the tasks investigated by \citep{Quoc_2015}, np-RNN performs better than IRNN or the corresponding scaled version of the IRNN, the iRNN, where the recurrent weight matrix is initialized to 0.01I.\\
The paper is organized as follows: In Section 2, we describe the simple RNN (sRNN) model and explain how the identity weight matrix initialization for hidden layer weights provide initial conditions to alleviate the long range problem in training sRNN. We then offer a dynamical systems perspective on the identity matrix weight initialization and present our proposal for weight initialization of hidden layer weight matrix to construct the np-RNN. In Section 3, we present an overview of the experiments to evaluate the performance of np-RNN, followed by results from our experiments on benchmark datasets in Section 4. The conclusion is presented in Section 5. 

\section{Dynamical systems perspective on RNNs}

\begin{figure}[h]
  \centering
    \includegraphics[scale=0.5]{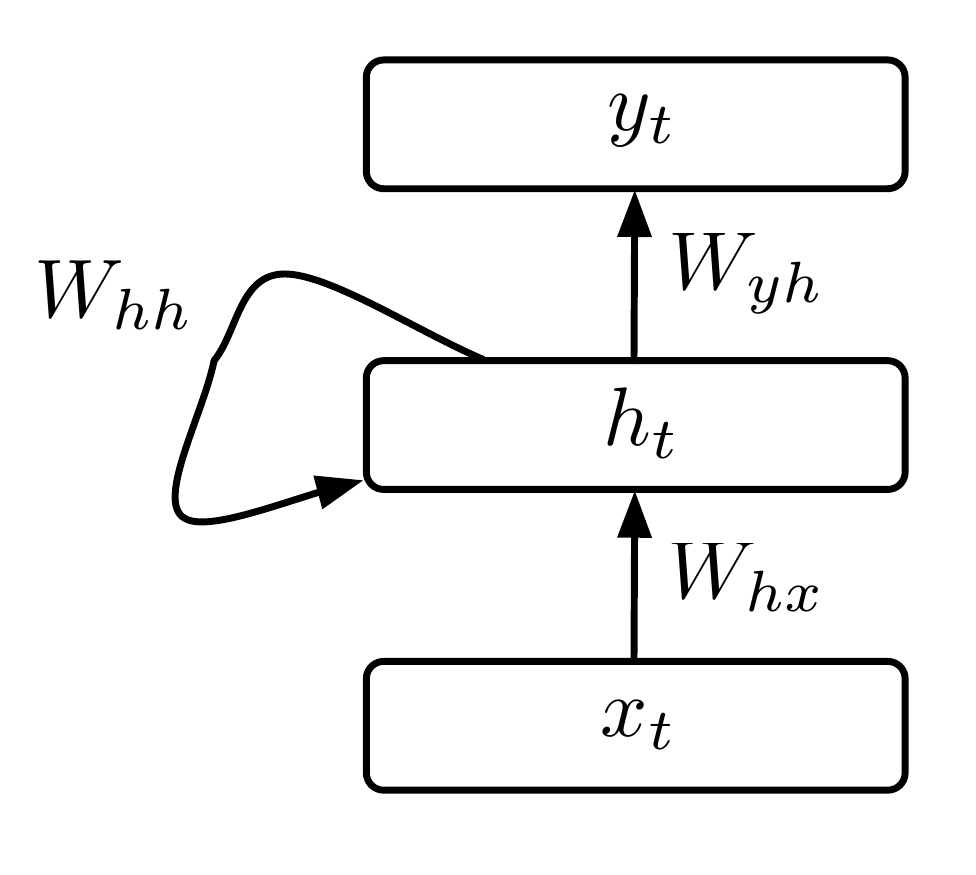}
  \caption{Schematic diagram of a simple RNN network  \label{Fig1}}
\end{figure}

In Figure \ref{Fig1} we show the schematic of a sRNN. The sRNN consists of an input layer, a hidden layer with recurrent connections and an output layer. Given an input sequence ${\bf X}=\{x_{0},x_{1},x_{2}\cdots x_{T}\}$, the sRNN will take the input $x_{t}\in \mathbb{R}^N$ and generate the prediction $y_t\in \mathbb{R}^C$ for the output sequence ${\bf Z}=\{z_{0},z_{1},z_{2}\cdots z_{T}\}$, where $z_{t}\in \mathbb{R}^{C}$ . In between the input and the output there is a hidden layer with $M$ units, which store information on the previous values ($t^{'}<t$) of the input sequence. More precisely, the sRNN takes $\bf{X}$ as input and generates an estimate ${\bf Y}$ of the output ${\bf Z}$ by iterating the equations
\bea
s_{t}&=&W_{hx}x_{t}+W_{hh}h_{t-1}+b_{h} \\
h_{t}&=&f(s_{t}) \\
o_{t}&=&W_{yh}h_{t}+b_{y}\\
y_{t}&=&g(o_{t})
\eea  
where $W_{hx}$, $W_{hh}$ and $W_{yh}$ are the weight matrices, $b_{h}$ and $b_{y}$
are the biases, $s_{t}\in \mathbb{R}^{M}$ and $o_{t}\in \mathbb{R}^{C}$ are inputs to the hidden layer and the output layer respectively and $f$ and $g$ are functions. For the purpose of this paper, $f$ is a ReLU and depending on the benchmark under investigations; $g$ is a linear function or a softmax function. In addition; all hidden layer nodes are assumed to be initialized to a fixed bias $b_{i}$ such that at t=0 $h_{0}=b_{i}$. Unless otherwise stated, for all experiments we set $b_{i}=0$. The objective function, $O(\theta)$ for sRNN with a single training pair (x,y) is defined as $O(\theta)=\sum_{t}L_{t}(z,y(\theta))$, where $\theta$ represents the set of parameters (weights and biases) in the sRNN. For regression problems, $L_{t}=\norm{(z_{t}-y_{t})^{2}}$ and for multi-class classification problems, $L_{t}=-\sum_{j}z_{tj}\log(y_{tj})$.

The IRNN model proposed by \citep{Quoc_2015} is a special case of sRNN with the following modifications: (i) the nonlinearity $f$ is ReLU and (ii) the hidden layer weights $W_{hh}$ are initialized with identity matrix and the bias terms are set to zero. In order to better understand how these modifications may enable RNN to overcome some of the long range problems, let us look at the gradient equation in the BPTT algorithm \citep{Pascuna_2013}:
\bea
\frac{\partial C}{\partial \theta}&=&\sum_{t\le T}\frac{\partial L_{t}}{\partial \theta} \\
&=&\sum_{t\le T}\frac{\partial L_{t}}{\partial h_{T}}\frac{\partial h_{T}}{\partial h_{t}}\frac{\partial^{+} h_{t}}{\partial \theta}
\eea
where,
\bea
\frac{\partial h_{T}}{\partial h_{t}}&=&\frac{\partial h_{T}}{\partial h_{T-1}}\frac{\partial h_{T-1}}{\partial h_{T-2}}\cdots \frac{\partial h_{t+1}}{\partial h_{t}}
\eea

Each Jacobian $\frac{\partial h_{t+1}}{\partial h_{t}}$ is a product of two matrices: (a) the recurrent weight matrix $W_{hh}$ and (b) diagonal matrix composed of the derivative of non-linearity, $f$, associated with the hidden nodes. In absence of any input, i.e., $x_{t}=0$, and with the choice of initial conditions for IRNN,the 2-Norm of each Jacobian in Equation 7 is identically one and the error gradients do not grow or decay exponential over time. 

In order to understand what the design of IRNNs means from a dynamical systems perspective, let us consider a simple example of sRNN with two hidden nodes with ReLU nonlinearity. We again assume that the sRNN receives zero input and all the biases are set to zero. We further assume that the recurrent weight matrix, $W_{hh}$, is positive definite. The dynamics of hidden nodes is then given by the following equation:
\bea
h^{'}_{t}&=&\Lambda h^{'}_{t-1} \quad\text{ if $h^{'}_{t-1}>0$} \nonumber \\
&=&0 \quad \text { ohterwise}
\eea
where $\Lambda=P^{-1}W_{hh}P$ is the diagonal matrix, $P$ is a matrix composed of eigenvectors of $W_{hh}$ and $h^{'}_{t}=P^{-1}h_{t}$.
\begin{figure}[h]
  \centering
    \includegraphics[scale=0.5]{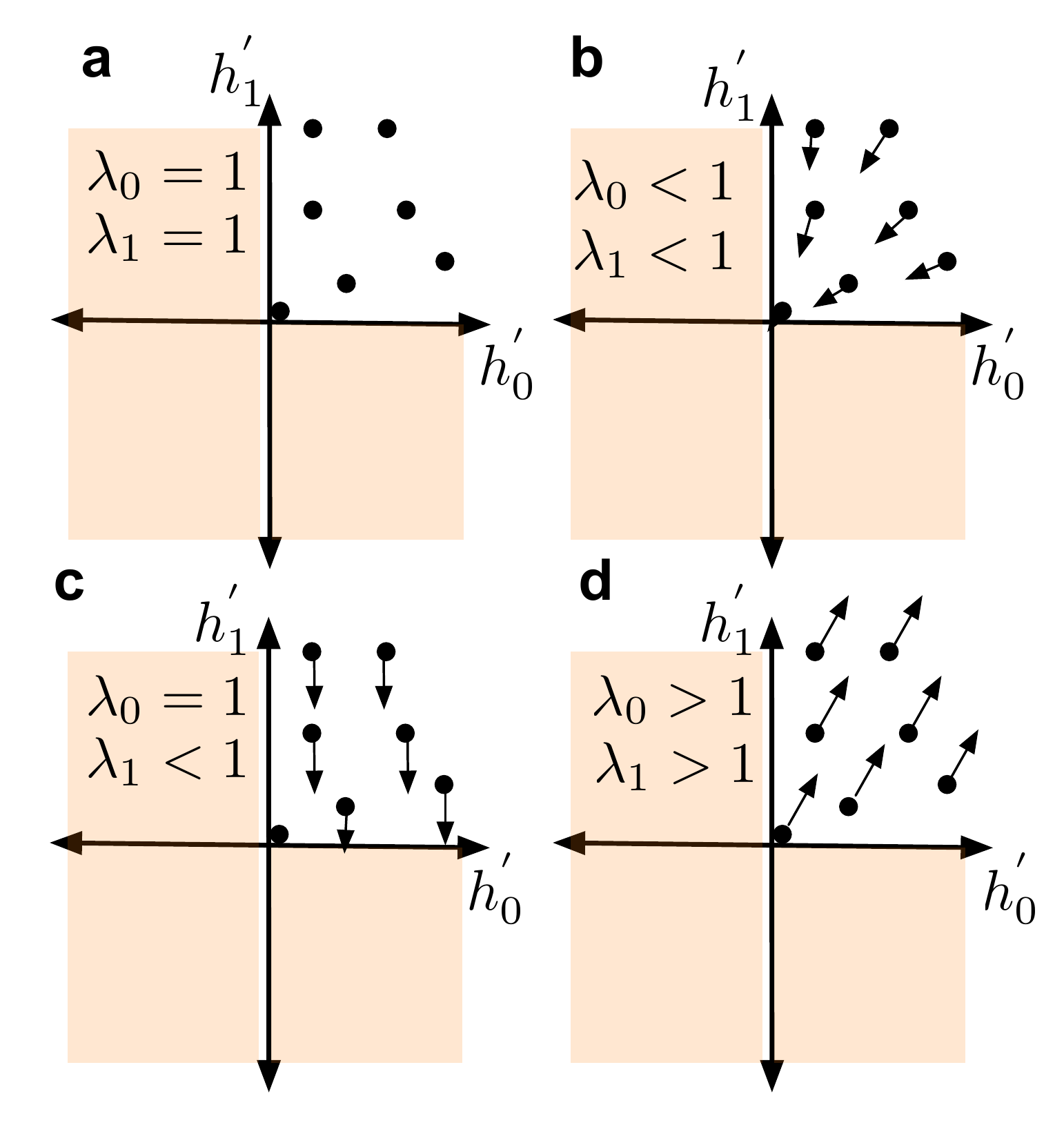}
  \caption{Schematic diagram of the phase space of a two-dimensional hidden node dynamical system. The regions in light orange correspond to the forbidden zone, given that the hidden nodes exhibit ReLU non-linearity (a) All the initial conditions are neutrally stable fixed points  (b) The dynamical systems has a global stable fixed point attractor at origin (c) The $h_{0}$ axis is the stable manifold of the dynamical system (d) The dynamical system has no stable fixed point and in absence of any perturbations, the trajectory will diverge to $\infty$ \label{Fig2}}
\end{figure}

In Figure \ref{Fig2}a-d, we show the schematic diagram of the phase-space of the hidden node dynamical system in Equation 8, dependent on the eigenvalues, $\lambda$, of the recurrent weight matrix \citep{Strogatz_2014}. The IRNN model corresponds to the case shown in Figure \ref{Fig2}a. In absence of any perturbations, every choice of initial conditions for the hidden nodes represents a fixed point that is neutrally stable. As a result, the evolution of the hidden node dynamics is strongly dependent on the input perturbations.  

The dynamics in \ref{Fig2}b, corresponds to the case when all eigenvalues are less than 1. In this case, the hidden node dynamical system has a single stable fixed point at the origin. In absence of any input the the hidden nodes will evolve to zero. Increasingly stronger input perturbations are required to prevent the hidden node dynamics from evolving to origin as the hidden nodes evolve closer to the stable fixed point. 

The case in \ref{Fig2}c corresponds to a situation where one eigenvalue is unity and the other eigenvalue is less than 1. In this case, the principal axis with eigenvalue of unity is the stable manifold. The trajectory of hidden nodes all evolve towards this manifold, eliminating the conditions for exploding gradients. The hidden node dynamics do not evolve to origin except for cases where the initial conditions are on one of the principal axes with eigenvalue less than unity, thereby decreasing the likelihood for the system to exhibit the problem of vanishing gradients. We hypothesize that recurrent weight matrix initialized to such a weight matrix offers the best opportunity for training sRNN without the long range problem of vanishing or exploding gradients. We conduct a series of experiments in support of the hypothesis. 

Finally, the case in Figure \ref{Fig2}d corresponds to the case when at least one eigenvalue is greater than 1. In this case, the hidden node dynamical system does not have any stable attractor and in absence of any input, the trajectory of hidden nodes will exponentially diverge to $\infty$ quickly resulting in the exploding gradient problem.

Based on the above observation, we propose the following equation for choosing the initial recurrent weight matrix, $W_{hh}$ and the corresponding RNN is referred to as the np-RNN. Assuming there are N hidden nodes, we propose the following, 
\bea
A&=&\frac{1}{N}\left<R^{T},R\right> \nonumber \\ 
e&=&\max(\lambda(A+I)) \nonumber \\
W_{hh}&=&\frac{I+A}{e}
\eea
where,$\left<\right>$ refers to dot product, $R$ is a standard normal matrix with values drawn i.i.d. from a Gaussian distribution with mean zero and unit variance, and $\lambda(X)$ is the set of all eigenvalues of matrix X. Equation 9 guarantees that $W_{hh}$ is a positive definite matrix with highest eigenvalue of unity and all the remainder eigenvalues less than 1. We refer to sRNN designed with this initial condition as the np-RNN. 

Our proposed weight initialization for RNN comprised of ReLU nonlinearity has some resemblance to the  initialization of recurrent weight matrix in the Echo-state network (ESN) \citep{Jaeger_2003, Palangi_2013}. Specifically, we also renormalize $W_{hh}$ with respect to the spectral radius of the positive definite matrix A. However, as opposed to ESN, we impose the constraint that $W_{hh}$ be positive definite in relation to our use of ReLU nonlinearity, which forces the hidden node dynamics to positive quadrant of the phase space. Relaxing the positive-definite constraint may result in a normalized initial weight matrix with complex eigenvalues. The dynamics of hidden node will then exhibit oscillations, and with ReLU nonlinearity, the hidden node dynamics will evolve to the stable fixed point at origin, which is not a desirable property for training RNNs. 

In the next Section, we present empirical evidence to support the requirement for positive-definite constraint on weight initialization for successfully training RNNs with ReLU nonlinearity. In addition, we also present empirical evidence demonstrating importance of ReLU nonlinearity as opposed to sigmoidal nonlinearity, which can by itself trigger vanishing gradient problem even with normalized positive definite weight initialization. 

We conclude this Section by noting that other than renormalization, the ESN does not impose any constraints on the choice of hidden node nonlinearity or the structure of the hidden node weight matrix. On the other hand, our proposed heuristics of normalized positive definite recurrent weight initialization and the choice of ReLU nonlinearity is necessary to successfully training RNNs.

\section{Experiments}
\begin{table*}[h]
\centering
\label{}
\begin{tabular}{|l|l|}
\hline
RNN Type & Description                                                                                                                                                                         \\ \hline
sRNN     & RNN with an input layer, single hidden layer and an output layer                                                                         \\ \hline
IRNN     & sRNN with recurrent weight matrix initialized to Identity matrix                                                                                                                    \\ \hline
iRNN     & sRNN with recurrent weight matrix initialized to 0.01 times Identity matrix                                                                                                         \\ \hline
nRNN   & \begin{tabular}[c]{@{}l@{}}sRNN with recurrent weight matrix initialized to a normalized matrix with all\\ except the highest eigenvalue less than 1\end{tabular} \\ \hline
np-RNN   & \begin{tabular}[c]{@{}l@{}}sRNN with recurrent weight matrix initialized to a normalized-positive definite matrix with all\\ except the highest eigenvalue less than 1\end{tabular} \\ \hline
np-tanhRNN   & \begin{tabular}[c]{@{}l@{}}np-RNN with tanh hidden node nonlinearity \end{tabular} \\ \hline
oRNN   & \begin{tabular}[c]{@{}l@{}}sRNN with recurrent weight matrix initialized to orthogonal random matrix \end{tabular} \\ \hline
gRNN     & sRNN with recurrent weight matrix initialized to a random Gaussian matrix                                                                                                           \\ \hline
\end{tabular}
\caption{Summary of various RNNs investigated in this work \label{Table0}}
\end{table*}

In the following experiments we compare np-RNN against np-RNN with tanh nonlinearity (np-tanhRNN),  IRNN, iRNN, RNNs that use ReLU and orthogonal random matrix initialization (oRNN), RNNs that use ReLU with random Gaussian initialization (gRNN) and finally, RNNs with normalized random Gaussian initialization (nRNN). nRNN resembles ESN in that the recurrent weight matrix initialization is done using a normalized random Gaussian matrix without the additional imposed constraint that the weight matrix be positive-definite.  oRNN is included in experiments because orthogonal weight matrix initialization is a commonly adopted method for initializing recurrent weight matrix for LSTMs. Finally, we also include experiments with np-tanhRNN, which use the same initialization scheme applied to np-RNN except the nonlinearity is tanh instead of ReLU. These are summarized in Table \ref{Table0}. 

For all our investigations, the RNNs were designed to comprise of one hidden layer with 100 hidden nodes. Except np-tanhRNN, all RNNs comprise ReLU nonlinearity. The values for the non-recurrent weight matrix $W_{hx}$ are drawn i.i.d. from a Gaussian distribution with zero mean and variance 1/N. Motivated by findings from \citep{Sussillo_2015}, we also scale $W_{hx}$ by a factor of $\alpha$, which has consistently boosted performance of our trained RNN relative to those that were trained without the $\alpha$ scaling factor. 
 \bea
\alpha&=&\sqrt{2}\exp\left(\frac{1.2}{\max(N,6)-2.4}\right)
\eea

The values for the non-recurrent weight matrix $W_{yh}$ are also drawn i.i.d. from a Gaussian distribution with zero mean with variance of 2/(fan$_{\text{in}}$+fan$_{\text{out}}$) \citep{Glorot_2010}. Unless otherwise stated, all RNNs are trained using stochastic gradient descent (SGD) with BPTT.  Except for the benchmark action recognition problem, we did not specifically tune any of the hyperparameters associated with SGD training via a grid search method. We consider the following benchmark datasets in our investigations:
\begin{enumerate}[label=\textbf{\arabic*}]
\item {\bf The Addition problem}\\
The addition problem is a toy benchmark used to evaluate the power of RNNs in learning long-term dependencies \citep{Hochreiter_1997}. The input to the RNN is a two-dimensional sequence $X=\{x_{0}(t),x_{1}(t)\}\rvert_{t=0}^{T}$. At each time step the input consists of a random signal $x_{0}\in [0,1]$ and a mask signal $x_{1}(t)$. The mask signal has all zeros except at two time steps when it has a value of 1. The RNN is to be trained to predict the sum of $x_{0}$ at these two time steps.

\item {\bf The multiplication problem}\\
The multiplication problem is another toy benchmark very similar to the addition problem. The difference is that instead of predicting the sum of $x_{0}$ at two time steps when the mask has a value of 1, the goal is to predict the product of  $x_{0}$ when the masking signal is unity.

\item {\bf MNIST classification with sequential presentation of pixels}\\
This is yet another challenging toy problem where the objective is to classify the MNIST digits when the 784 pixels are presented sequentially to the recurrent net. We follow the procedure described in \citep{Quoc_2015} and the RNN reads one pixel at a time in a scanline order, starting at the top left corner of the MNIST image and ending at the bottom right corner of the image. The RNN is asked to predict the label for the input image after reading all the 784 pixel values resulting in a large long range dependency problem.

\item{\bf Action recognition benchmark}\\
We consider one real world example of action recognition task based on the standard UCF-101 benchmark \citep{Soomro_2012}. The task is to classify a given video clip of length varying from 2 s to $>$10 s into one of the 101 action classes. The benchmark contains a total of 13320 video clips divided into 25 groups. Each group contains video clips from all 101 action classes. Video clips of a given action in a given group share some commonalities. Usually the performance of a given classifier on this data set is evaluated using three splits of the training/testing data. For our experiments we report results for the split one, with training data made up of video clips from groups 8-24 and the test data made up of videos derived from groups 1-7.
\end{enumerate}
\section{Results}
\subsection{Addition Problem}
As suggested in \citep{Quoc_2015}, a basic baseline solution would be to always predict 1 as the output regardless of the inputs. This will produce a Mean Squared Error (MSE) of approximately 0.166. To solve this problem the recurrent net has to remember the two relevant numbers and their sum while ignoring the irrelevant numbers. The task gets difficult with increasing length $T$ of the sequence.

In order to evaluate the RNN networks (iRNN,IRNN,n-RNN, np-RNN,np-tanhRNN,oRNN,gRNN) we generate 100,000 training examples and 10,000 test examples. For all the reported results, we trained the network using SGD-BPTT optimization. The learning rate was fixed at 0.001 and the gradient clip parameter was set to 10. The batchsize was set to 1 and all the networks were trained for 10 epochs.  Following from \citep{Quoc_2015}, we began by evaluating the performance of RNNs on the addition benchmark for sequences of length $T=\{150,200,300,400\}$. 
\begin{figure}[h]
  \centering
    \includegraphics[scale=0.5]{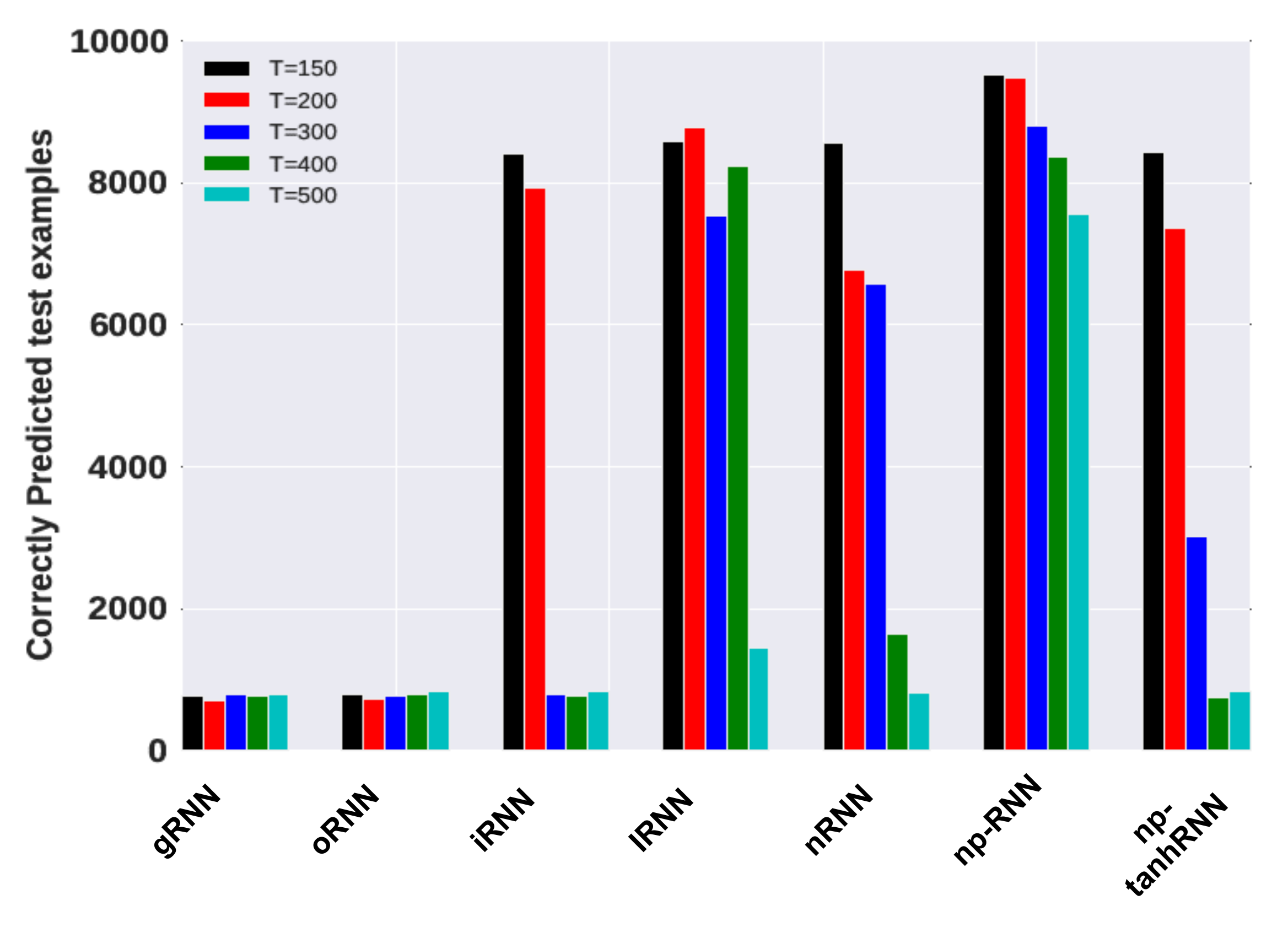}
  \caption{ Bar chart of correctly predicted test examples for the addition problem \label{Fig3}}
\end{figure}

The results are summarized in Figure \ref{Fig3}. We plot the total number of correctly classified sequences in the test data, defined as the sequences for which the absolute prediction error is less than 0.04 \citep{Martens_2011}. For sequences up to $T=300$, our results for gRNN and IRNN match those obtained by \citep{Quoc_2015}. We also note that the np-RNN performs at par or better than the other RNNs for $T<=300$. However, for sequence of length $T=400$, contrary to findings in the \citep{Quoc_2015}, we were able to train the IRNN as well as the np-RNN. We therefore also evaluated the performance of the RNNs for sequence of length $T=500$. Here we find that np-RNN can still be successfully trained whereas the training is challenging for the IRNN and all other RNNs.   

We also evaluated the performance of oRNN, nRNN and np-tanhRNN on the Addition problem. Similar to gRNN, oRNN failed to train on the Addition problem for all the sequence lengths that we investigated. The nRNN and the np-tanhRNN, on the other hand, performed relatively well on the Addition problem for short input sequences ($T=150,200,300$). We conjecture that both np-tanhRNN and nRNN suffer from the vanishing gradient problem as the length of input sequence is increased. We believe that the vanishing gradient problem for np-tanhRNN is primarily caused by the fact that $\frac{\partial h_{t+1}}{\partial h_{t}}<1$. The vanishing gradient problem with nRNN is related to the fact that the the  recurrent weight matrix is initialized with a normalized random matrix with imaginary eigenvalues, which result in oscillatory hidden node dynamics and with ReLU nonlinearity, the network can fairly quickly evolve to a fixed point at the origin. Systematic analysis of this conjecture for failure of nRNN remains a topic of future investigation.

\subsection{Multiplication problem}
In this case a baseline solution would be to always predict the two numbers to be 0.5 such that the product is 0.25. This will produce a Mean Squared Error (MSE) of around 0.2025. Again the goal is to train RNN to produce MSE much lower than 0.2025. We follow \citep{Martens_2011} and evaluate the performance of RNNs for sequence of length $T=\{50,100,200\}$. Since gRNN and oRNN could not be trained on the addition benchmark for any sequence of length $T\ge 150$ , we did not pursue training of gRNN for the multiplication problem. 
\begin{figure}[h]
  \centering
    \includegraphics[scale=0.5]{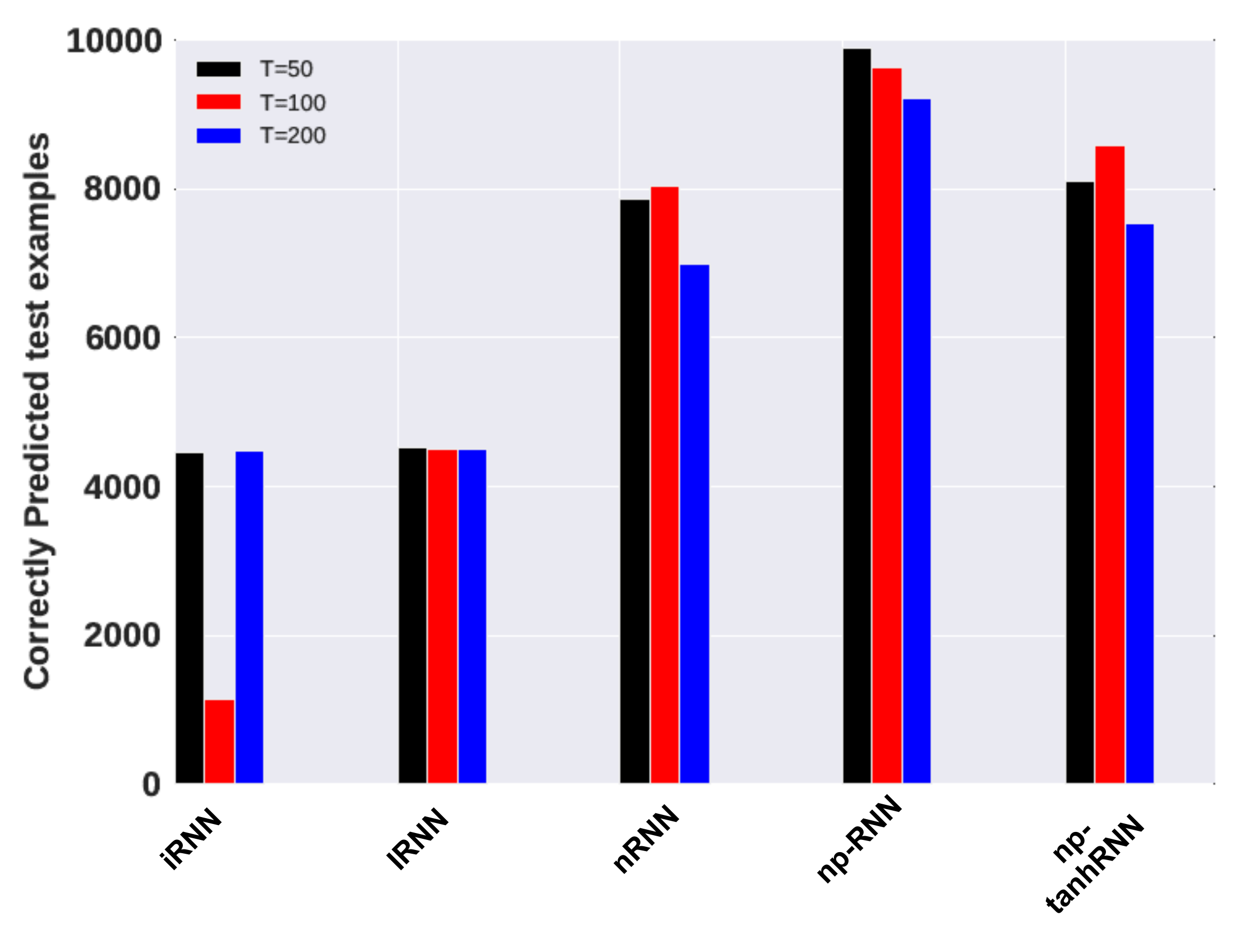}
  \caption{ Bar chart of correctly predicted test examples for the multiplication problem \label{Fig4}}
\end{figure}
We again generated 100,000 training examples and 10,000 test examples and trained all the networks using SGD-BPTT optimization. 
The training continued for 100 epochs with batch size of 16. The learning rate initially was set at 0.0002 and then cooled twice by a factor of 10 in equal intervals. The results are summarized in Figure \ref{Fig4}. Similar to the addition benchmark, we plot the total number of correctly classified sequences in the test data, defined as the sequences for which the absolute prediction error is less than 0.04. We see that neither iRNN, nor IRNN can correctly classify greater than 50 \% of test sequences of lengths $T$ investigated. However, np-RNN produces classification accuracy $> 90$ \% for all sequences investigated. We also note that nRNN and np-tanhRNN did fairly well on all sequence lengths investigated for the multiplication problem. These findings are similar to those obtained for the Addition problem. np-RNN in general outperforms both nRNN and np-tanhRNN due to its ability to better handle the vanishing gradient problem by initially forcing the gradients $\frac{\partial h_{t+1}}{\partial h_{t}}=1$ and restricting the hidden node dynamics to the positive quadrant of the phase space.

\subsection{MNIST classification with sequential presentation of pixels}
This is another challenging toy problem where the goal is to classify MNIST digits \citep{Lecun_MNist} when the 784 pixels are presented sequentially to the RNN. The network is asked to predict the digit after all the 784 pixels are presented to the RNN serially one pixel at a time.  
\begin{figure}[h]
  \centering
    \includegraphics[scale=0.5]{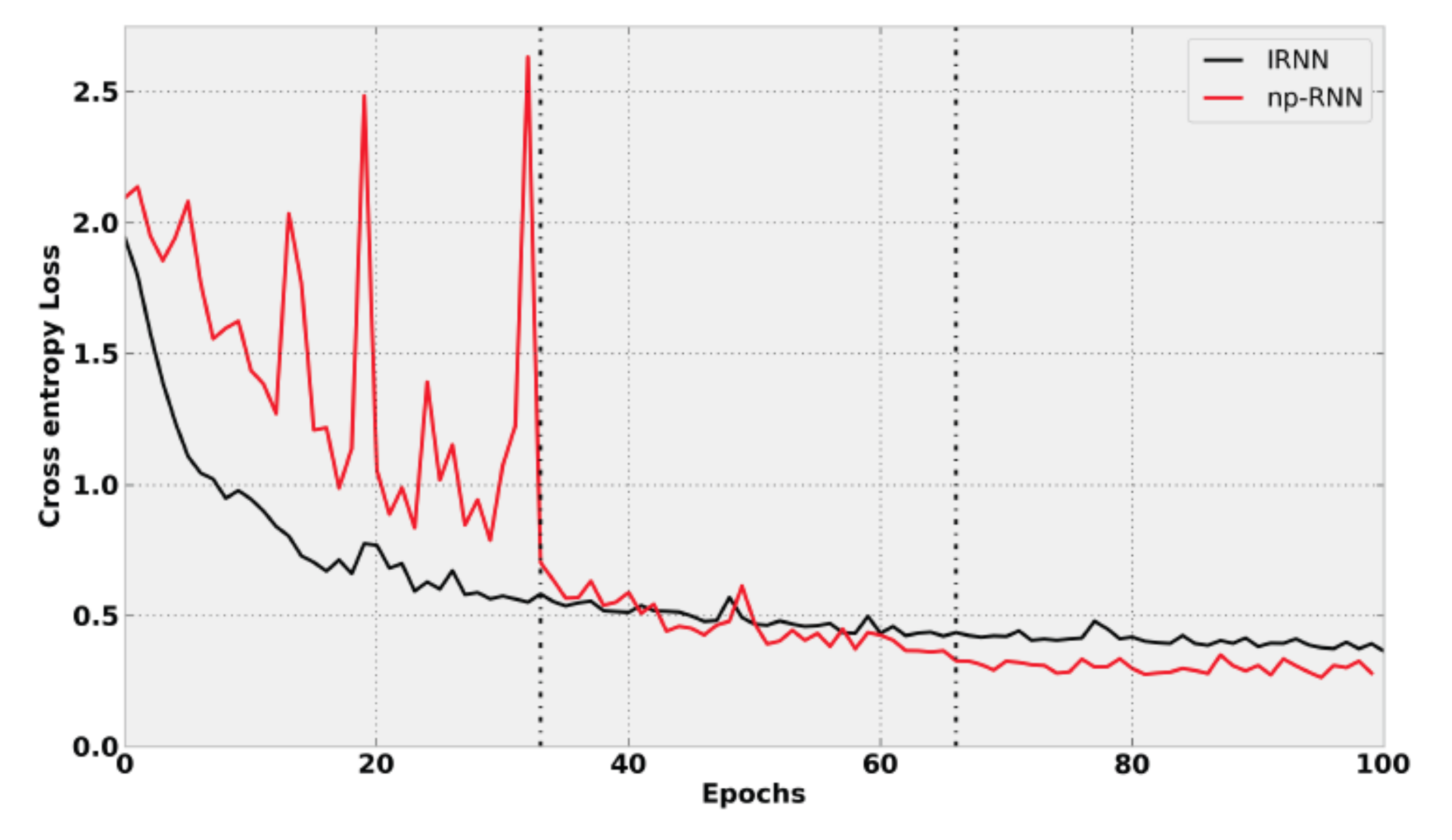}
  \caption{ The loss function on the test data plotted as function of the number of epochs. The dotted lines correspond to the times when the learning rate was cooled by a factor of 10. Note the high variability in the loss function for np-RNN in the first phase of learning is because of large choice for initial learning rate relative to those used in training the IRNN. \label{Fig5}}
\end{figure}
We began by first attempting to reproduce the findings from \citep{Quoc_2015}. However using the SGD-BPTT optimization, we were unable to train IRNN on MNIST using the set of parameters reported in \citep{Quoc_2015}. Using RMSprop \citep{RMSProp}, however, we were successful in training IRNN with learning parameter of $10^{-6}$. The np-RNN was trained using SGD-BPTT optimization with cooling.  We trained both IRNN and np-RNN network for 100 epochs with batch size of 16, corresponding to about 375000 iterations. The np-RNN was trained for 33 epochs with initial learning rate of $10^{-4}$, followed by training for another 33 epochs with learning rate cooled by factor of 10 and then again trained unto 100 epochs by cooling learning rate by another factor of 10. In Figure \ref{Fig5}, we plot the evolution of the cost function on the test batch as a function of the number of epochs.
The resulting accuracy is reported in Table \ref{Table1}. We note that while the IRNN produces accuracy of about 83 \%, the np-RNN produces an accuracy of about 92 \%, which is inline with the results reported by \citep{Quoc_2015}. The accuracy grows to 96.8\% on training the np-RNN for 500 epochs,which corresponds to 937500 epochs, while for IRNN trained with the given parameters, the accuracy grows to about 93\%. Based on our findings on the performance of IRNN and np-RNN on the action recognition benchmark (see Section 4.4) and our inability to reproduce the results reported in \citep{Quoc_2015}, we conclude that IRNN is extremely sensitive to the choice of the training protocol and the hyper parameters of the network. On the other hand, the observation that np-RNN performs as well as published results for this toy problem is encouraging.

\begin{table}[h]
\centering
\label{}
\begin{tabular}{|l|l|l|l|}
\hline
RNN Type & Training Method & \begin{tabular}[c]{@{}l@{}}Accuracy\\ 100 epochs\end{tabular} & \begin{tabular}[c]{@{}l@{}}Accuracy\\ 500 epochs\end{tabular} \\ \hline
IRNN     & RMSProp         & 83 \%                                                         & 93 \%                                                         \\ \hline
np-RNN   & SGD-BPTT        & 92 \%                                                         & 96.8 \%                                                       \\ \hline
\end{tabular}
\caption{Performance on MNIST digit classification\label{Table1}}
\end{table}

\subsection{Action recognition benchmark}
We also benchmarked iRNN, IRNN, np-RNN and the LSTMs on the UCF-101 action recognition dataset \citep{Soomro_2012}.   We followed the approach proposed in \citep{Ng_2015} and trained the RNNs on the sequence of features generated by passing the sequence of video frames through a pre-trained Convolutional Neural Network (CNN).  Motivated by the recent findings of \citep{Cees_2015}, we use a CNN trained on 15000 Imagenet object categories as the feature generator. The CNN architecture was modeled after the Googlenet-CNN \citep{Googlenet} and the features were extracted from the last fully-connected layer of the Googlenet-CNN.

The UCF101 benchmark contains 13,320 video clips, divided into 25 groups. The groups are divided into 3 splits of training and testing for evaluation. As the training data is very small, we did not fine tune the CNN, but rather only focused on training the RNNs using raw features generated by a pre-trained feature learner. Furthermore to avoid overfitting, we introduced dropout between the hidden and the output layer of the RNN.

Our implementation of LSTM was standard including the forget gate. For all the 4 RNNs, we did a grid search over learning rates $\l=\{10^{-5}, 5\times 10^{-5},10^{-4}, 5\times 10^{-4},10^{-3},10^{-2}\}$ and the dropout parameter $d=\{0.5,0.7,0.9\}$. The training protocol was: RMSprop for optimization, batchsize of 256, training for a total of 70 epochs, with training for the first 25 epochs with initial learning rate, followed by training for additional 30 epochs with learning rate cooled by factor of 10 and final phase training for additional 15 epochs by cooling the learning rate by another factor of 10. For all our investigations, we used the split one of the training/testing data comprised of video clips from groups 8 through 25 in the training set and video clips from group 1 through 7 in the test set. In order to search over the space of hyperparameters $\{l,d\}$, we further split the split one training data into training and validation data set. Data from groups 8 through 24 was used to train all the RNNs and data from group 25 was used for validation. 
\begin{figure*}[h]
  \centering
    \includegraphics[scale=0.5]{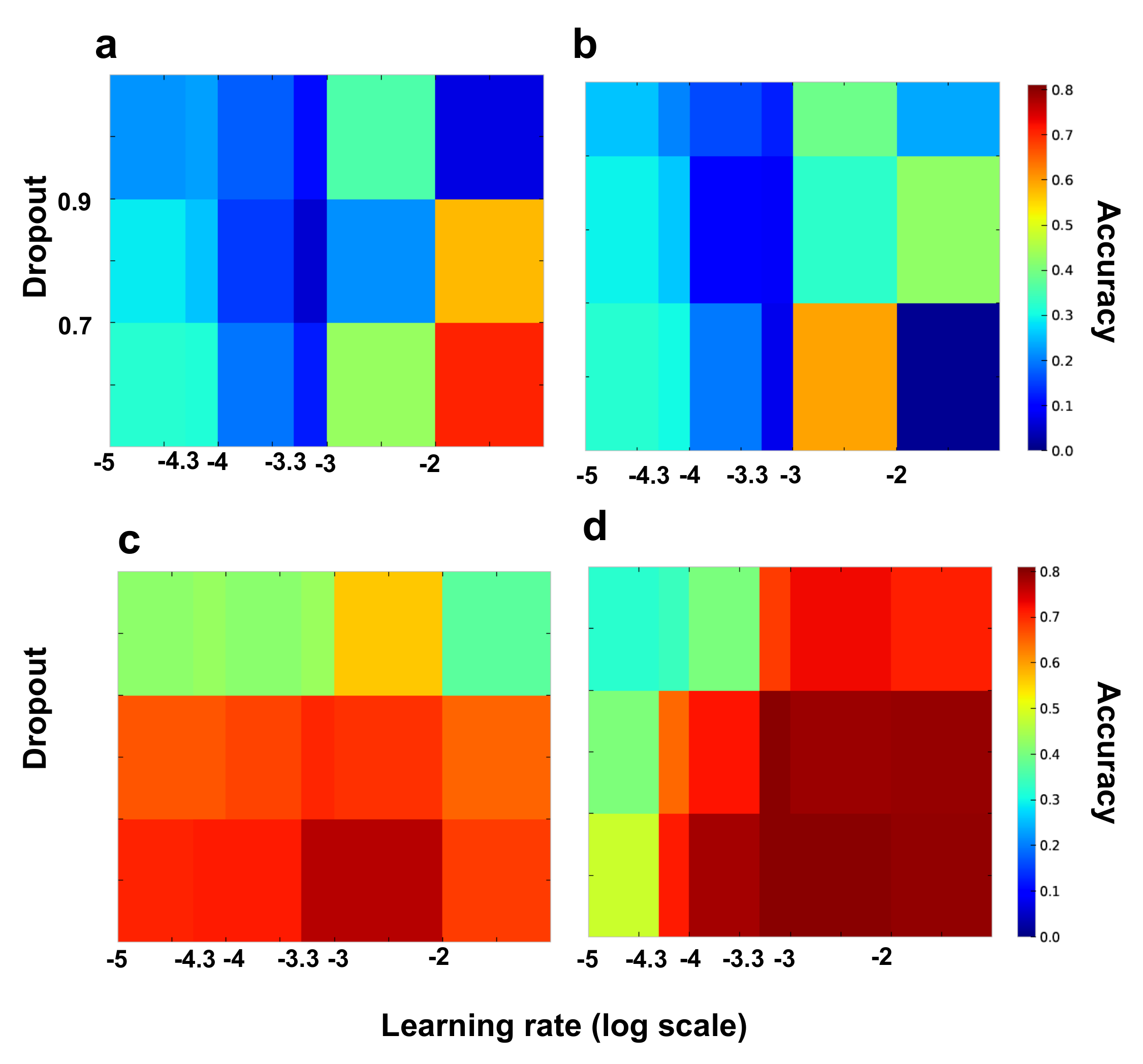}
  \caption{ Plot of accuracy on the validation data as a function of the grid search parameters: Learning rate and dropout for (a) IRNN (b) iRNN (c) np-RNN and (d) LSTM \label{Fig6}}
\end{figure*}
In Table \ref{Table2} we report the score on the split one test data using RNN with best validation score over the grid search. The scores for np-RNN are closer to LSTM than IRNN and are comparable to the published state-of-the-art for split one of training/test data for the UCF101 benchmark \citep{Ng_2015,Cees_2015}. iRNN performed the worst for the set of hyper-parameters investigated. Note that while the LSTM produces the best scores for this benchmark, LSTM is  four times computationally complex (in terms of number of free trainable parameters) relative to the sRNN \citep{Klaus_2015}.

In Figure \ref{Fig6}, we plot the validation scores for the 3 RNNs as function of the grid search parameters. It is clear from the Figure that the performance of both iRNN and IRNN are strongly dependent on the right choice of hyperparameters, which is not the case for np-RNN or the LSTM. 
\begin{table}[t]
\centering
\begin{tabular}{| c | L{2.5cm} | C{2.2cm} | R{2cm} |}
\hline
RNN Type & Training Method                                                    & \begin{tabular}[c]{@{}l@{}}Accuracy \\validation \end{tabular} & \begin{tabular}[c]{@{}l@{}}Accuracy\\ test\end{tabular} \\ \hline
iRNN     & \begin{tabular}[c]{@{}l@{}}RMSProp \\ (l=$10^{-3}$; d=0.5)\end{tabular} & 59 \%                                                                           & 56.6 \%                                                                  \\ \hline
IRNN     & \begin{tabular}[c]{@{}l@{}}RMSProp \\ (l=$10^{-2}$; d=0.5)\end{tabular} & 71 \%                                                                           & 67 \%                                                                  \\ \hline
np-RNN   & \begin{tabular}[c]{@{}l@{}}RMSProp\\ (l=$10^{-3}$; d=0.5)\end{tabular} & 78.6 \%                                                                         & 75.2 \%                                                                \\ \hline
LSTM     & \begin{tabular}[c]{@{}l@{}}RMSProp\\ (l=$10^{-3}$; d=0.5)\end{tabular} & 80.3 \%                                                                         & 78.5 \%                                                                \\ \hline
\end{tabular}
\caption{Performance of RNNs on split one of UCF101 benchmark\label{Table2}}
\end{table}
\section{Conclusion}
We offer a dynamical systems perspective on the Identity weight initialization for the recurrent weight matrix for RNNs with hidden nodes made up of ReLUs. We hypothesize that the sensitivity of hidden nodes to input perturbations resulting from the Identity weight matrix initialization, may make IRNN sensitive to the choice of hyperparameters for successful training. We offer an alternative weight initialization strategy, the normalized positive-definite weight matrix, that attempts to reduce the sensitivity of the hidden nodes to input perturbations by collapsing the dynamics to a one-dimensional manifold. We compare the performance of IRNN to np-RNN on several toy examples and also one real world action recognition benchmark and demonstrate that np-RNN either performs better or is comparable to IRNN on these benchmarks.

\subsubsection*{Acknowledgments}
We acknowledge valuable feedback from the anonymous reviewers at ICLR and our colleagues, Daniel Fontijne and Anthony Sarah at Qualcomm Research. We also acknowledge kind assistance from Cees Snoek and Mihir Jain in sharing their Googlenet-CNN trained on the entire Imagenet database.

\bibliography{Reference}
\bibliographystyle{iclr2016_conference}

\end{document}